\documentclass{amia}
\usepackage{lipsum} 
\usepackage{multirow}
\usepackage{amssymb}
\usepackage{amsmath}
\usepackage{color}
\usepackage{booktabs}
\usepackage{xcolor}
\usepackage{longtable,natbib}
\usepackage{wrapfig}
\usepackage{float}
\usepackage{soul}
\usepackage{framed}
\colorlet{shadecolor}{blue!20}

\usepackage{tikz}
\newcommand{\casebox}[1]{\tikz[baseline=(X.base)]\node [draw=red!50,fill=pink!15,rectangle,inner sep=1pt, rounded corners=2pt] (X) {#1};}

\def\Algnameabbr{LLM-PTM}

\begin{document}


\title{Large Language Models for Healthcare Data Augmentation: An Example on Patient-Trial Matching}

\author{Jiayi Yuan$^1$, Ruixiang Tang$^1$, Xiaoqian Jiang, PhD$^2$, Xia Hu, PhD$^1$ }

\institutes{
    $^1$ Rice University, Houston, TX; $^2$University of Texas Health Science Center, Houston, TX.
}

\maketitle

\section*{Abstract}


\textit{The process of matching patients with suitable clinical trials is essential for advancing medical research and providing optimal care. However, current approaches face challenges such as data standardization, ethical considerations, and a lack of interoperability between Electronic Health Records (EHRs) and clinical trial criteria. In this paper, we explore the potential of large language models (LLMs) to address these challenges by leveraging their advanced natural language generation capabilities to improve compatibility between EHRs and clinical trial descriptions. We propose an innovative privacy-aware data augmentation approach for LLM-based patient-trial matching (LLM-PTM), which balances the benefits of LLMs while ensuring the security and confidentiality of sensitive patient data. Our experiments demonstrate a 7.32\% average improvement in performance using the proposed LLM-PTM method, and the generalizability to new data is improved by 12.12\%. Additionally, we present case studies to further illustrate the effectiveness of our approach and provide a deeper understanding of its underlying principles.}

\section{Introduction}
\vspace{-0.5em}


Identifying suitable clinical trials for patients is crucial for the reliable assessment and examination of medical treatments. This process not only enables patients to obtain the best possible care but also supports advancements in medical research. Models that facilitate clinical trial matching compare patient profiles with the eligibility requirements of ongoing studies to pinpoint potential matches \cite{Mattson1994-nk}. Nonetheless, discovering the ideal clinical trial for a patient can be a complex and lengthy procedure. This is where advanced clinical trial matching methods come into play. By employing artificial intelligence (AI) or specific algorithms, researchers can pair patients with fitting clinical trials \cite{Hassanzadeh2020-is,Alexander2020-fr}, a task that is growing increasingly vital as the number of studies continues to expand. Please refer to our preliminary section for more details.

In spite of the potential benefits, integrating AI into clinical trial processes encounters various obstacles, such as issues pertaining to data availability, standardization, and ethical considerations \cite{Bhatt2021-od}. A significant challenge lies in the discordance between the ontology and terminology utilized in Electronic Health Records (EHRs) and those implemented in clinical trial inclusion and exclusion criteria. Although recent research has produced solutions to address these concerns through black-boxed embedding matching \cite{gao2020compose, zhang2020deepenroll}, the efficacy of AI-driven clinical trial matching services may still be impeded by the difficulty in interoperating the information from the two disparate sources.


The advent of large language models (LLM) \cite{openai2023gpt4} presents an opportunity to enhance the compatibility between EHRs and clinical trial descriptions, promoting more accurate patient-trial matching. By capitalizing on their sophisticated natural language processing abilities, these models can efficiently decipher, comprehend, and harmonize the diverse terminologies and ontologies present in both EHRs and clinical trial inclusion and exclusion criteria. This enhanced interoperability not only streamlines the matching process but also ensures greater accuracy in identifying suitable trials for patients. Consequently, large language models hold the potential to transform clinical trial matching, enabling better patient outcomes and contributing to more efficient medical research.

The foremost challenge when implementing LLMs for clinical trial matching lies in managing privacy concerns that arise due to the handling of sensitive patient data, which may include personal and health-related information. Ensuring the security and confidentiality of this data is crucial to maintain patient trust and adhere to legal and ethical standards. To overcome this challenge, we propose an innovative data augmentation that prioritizes data privacy while maintaining the benefits offered by LLMs. During the actual implementation, instead of feeding original patient data directly into LLMs, we use desensitized patient data as a prompt to guide the LLM in the augmentation process of the trial data. Through comprehensive experimentation, our proposed \Algnameabbr{} demonstrates an average improvement in performance by $7.32\%$, it also improves generalizability by $12.12\%$. Moreover, we conduct case studies to elucidate the efficacy of the method and provide a broader insight into its underlying principles.




\section{Preliminary}
\vspace{-0.5em}

\subsection{Patient Trial Matching}
\vspace{-0.5em}

\begin{figure*}[t!]
\centerline{\includegraphics[width=0.9\textwidth]{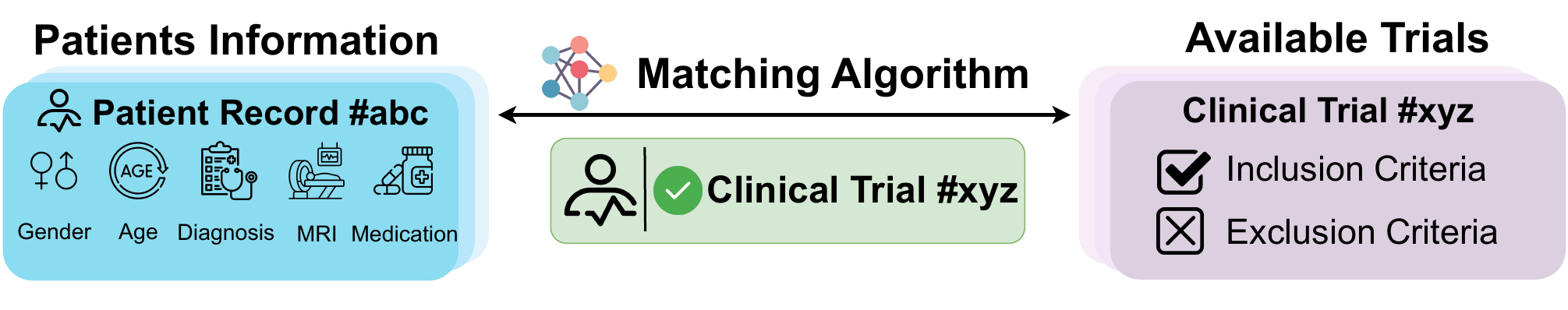}}
\caption{An illustration of patient-trial matching.}
\label{fig:illustration}
\end{figure*}

In the realm of clinical research, clinical trials serve as the solely established methodology for the development of novel disease treatments. However, these trials frequently encounter challenges such as costly, imprecise, and inadequate patient recruitment. A substantial number of trials grapple with obtaining the necessary patient population, with $50\%$ experiencing delays due to recruitment issues, while others fail to secure a sufficient number of participants to initiate the trial \cite{hargreaves2016clinical}. The advent of automated patient-trial matching offers a promising avenue for optimizing the trial recruitment process. The crux of this approach lies in identifying eligible patients for clinical trials based on their longitudinal electronic health records (EHR) and trial eligibility criteria (EC), which encompass both inclusion and exclusion criteria as depicted in Fig.~\ref{fig:illustration}. The problem can be framed as a classification problem. Given input consisting of a patient's complete Electronic Health Record (EHR) data and a single trial's eligibility criteria (EC), the output can be classified as either match, mismatch, or unknown. A match between a patient and a trial occurs if and only if the patient satisfies all of the trial's ECs. Mathematically, let $P$ represent the patient's EHR data and $T=[t_1,...,t_n]$ represent the trial's eligibility criteria. The matching function $m(P,t)=\{match, mismatch, unknown\}$ and $M(P,T)=\{match, mismatch\}$.

\subsection{Data Augmentation}
\vspace{-0.5em}

Text data augmentation, involving diverse transformations, is widely utilized to improve model training for text classification tasks within the domain of natural language processing (NLP). Contemporary data augmentation techniques in NLP function at distinct levels of granularity, including characters, words, sentences, and documents. The objective of data augmentation is to generate intelligible and varied additional instances that uphold semantic congruity. In the context of patient-trial matching, data augmentation assumes particular importance due to the limited richness of the available training data. The implementation of an effective augmentation method can facilitate the generation of a more varied text dataset, allowing machine learning models to better capture the intricacies of patient and eligibility criteria information. Consequently, this leads to enhanced training of classifiers, resulting in more accurate and reliable outcomes for patient-trial matching tasks.


\subsection{Data Privacy in Healthcare}
\vspace{-0.5em}

The potential of big data to revolutionize healthcare is acknowledged \cite{price2019privacy}, but privacy concerns remain. Privacy issues lead to both consequentialist and deontological concerns in data usage \cite{price2019privacy}. Restricting access to patient data can impede data-driven innovation \cite{price2017drug}, while data deidentification complicates linking patient data from different sources \cite{eisenberg2017promoting, evans2018big}. Alternative privacy-preserving approaches, such as pseudonymized data or differential privacy techniques, can be applied in certain contexts \cite{beaulieu2019privacy, dwork2014algorithmic, moussa2017differential}. Privacy audits and security standards can also help, with data holders acting as stewards rather than privacy-agnostic intermediaries. Yet, a privacy-innovation tradeoff may still persist in many situations.
In the context patient-trial matching, privacy concerns also arise due to the inherent nature of big data. This study explores privacy-conscious augmentation techniques that interact with open-source components, aiming to improve match identification accuracy while preventing the leakage of private-source data.


\section{Methodology}
\vspace{-0.5em}


\subsection{Problem Setting}
\vspace{-0.5em}

As previously elucidated, the process of patient-trial matching entails the identification of suitable patients for a specific clinical trial based on their Electronic Health Records (EHR). These records contain comprehensive medical information about the patient, while clinical trials are characterized by detailed descriptions, eligibility criteria, and other pertinent data. In the following section, we will systematically introduce the problem setting.

\textbf{Input 1: Patient Records}. We extract patient records from raw clinical documentation and subsequently transform them into a structured tabular format. We represent patient records by the symbol $P$ and formally define it as: 
$P = [d_1, d_2, \ldots, $ $d_{n_d}, m_1, m_2, \ldots, m_{n_m}, p_1, p_2, \ldots, p_{n_p}]$. 
Here, $d_i$ refers to a diagnosis belonging to the set $\mathcal{D}$ of diagnoses, $m_i$ represents a medication from the set $\mathcal{M}$ of medications, and $p_i$ denotes a procedure within the set $\mathcal{P}$ of procedures. All elements, i.e., diagnoses, medications, and procedures, are character strings. The quantities $n_d, n_m,$ and $n_p$ are the total numbers of diagnoses, medications, and procedures in the electronic health record (EHR) categories, respectively.

\textbf{Input 2: Clinical Trials}. In our problem setting, we focus on the criteria of clinical trials. Analogous to the extraction of patient notes, we retrieve eligibility criteria from unprocessed patient trial documents. Let us denote clinical trials by the symbol $T$ and express it as:
    $T = [i_1, i_2, \ldots, i_{n_i}, e_1, e_2, \ldots, e_{n_e}]$.
In this representation, $i_i$ and $e_i$ symbolize the inclusion and exclusion criteria, respectively, both of which are character strings. The variables $n_i$ and $n_e$ indicate the total number of inclusion and exclusion criteria, correspondingly.

\textbf{Task 1: Patient-Criteria Matching}. Given a patient's visit records $P$ and a collection of inclusion or exclusion criteria, we frame the patient-criteria matching task as a multi-class classification problem. The objective is to categorize the matching outcomes between patients and Eligibility Criteria (ECs) into three distinct classes: ``match'', ``mismatch'', and ``unknown'', according to the similarity between patient records and trial criteria. We can represent this as:
$\hat{y}(c, P) \in \{match,\ mismatch,\ unknown\}$, where $c \in T.$

\textbf{Task 2: Patient-Trial Matching}. Given a patient's visit records $P$ and a clinical trial $C$ consisting of a collection of inclusion and exclusion criteria, we assert that a patient and a trial constitute a match only if the patient satisfies all inclusion criteria and contradicts all exclusion criteria in the trial. We can represent this condition as: $ Match(C, P) \Leftrightarrow (\forall i \in \{i_1, \ldots, i_{n_i}\}, : \hat{y}(i, P) = match) \wedge (\forall e \in \{e_1, \ldots, e_{n_e}\}, : \hat{y}(e, P) = match) $.

\subsection{Proposed Pipeline}
\vspace{-0.5em}

\textbf{Trial Eligibility Criteria Augmentation}

As previously discussed, the acquisition of comprehensive and high-quality data in patient-trial matching presents significant challenges, including considerable expenses and potential privacy infringement. Recognizing the necessity for data augmentation in this context, we have introduced \Algnameabbr{}. In this work, we put forth a data augmentation technique utilizing Language Models (LLMs) to create supplementary data points while preserving the semantic coherence of the original trial's inclusion (i) and exclusion (e) criteria. We first employ Chain-of-Thought to direct the LLMs to gradually generate the prompts. These prompts incorporate our requirement that the output data be more comprehensible to machine learning models, while preserving the exact semantic content. Subsequently, we use desensitized patient data, clinical trial data, and the previously generated prompts to execute a data augmentation process that preserves privacy.

\begin{figure*}[t!]
\centerline{\includegraphics[width=1\textwidth]{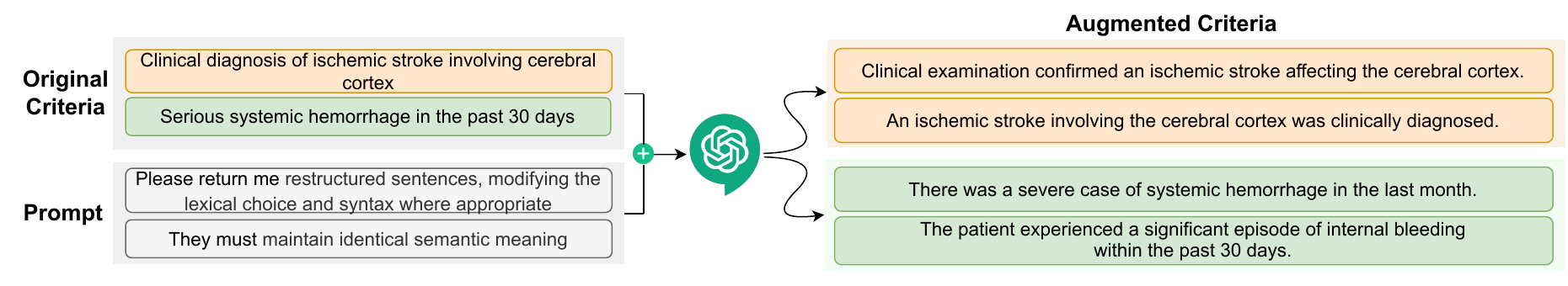}}
\caption{Illustration of \Algnameabbr{} augmented criteria.}
\label{fig:augment}
\end{figure*}

The illustration and examples of the augmentation part of \Algnameabbr{} can be found in Fig.~\ref{fig:augment}. Given the criteria of a clinical trial: $T = [i_1, i_2, \ldots, i_{n_i}, e_1, e_2, \ldots, e_{n_e}]$, we aim to utilize an LLM to generate a set of augmented data points $\mathcal{T}$ that adhere to these constraints. Formally, for each criteria $i_k\ \text{and}\ e_l \in T$, we construct input strings $i_k'$ and $e_l'$ as follows:
$
    i_k' = o \oplus i_k$,
    $e_l' = o \oplus e_l
$
where $o$ is the designed prompt, $\oplus$ denotes the concatenation operation. Subsequently, we feed the input strings $i_k'$ and $e_l'$ into the LLM to generate a set of augmented data points $A_{i_k}$ and $A_{e_l}$, respectively. Regard the LLM as a function $LLM()$: $ A_{i_k} = LLM(i_k')$, $A_{e_l} = LLM(e_l')$. The final augmented trial dataset $\mathcal{T}$ can then be represented as:

\vspace{-1em}
\begin{equation}
\mathcal{T} = \bigcup_{k=1}^{n} A_{i_k} \cup \bigcup_{l=1}^{m} A_{e_l}
\end{equation}

\textbf{Patient and Criteria Embedding}

After getting preprocessed text data, we then do embedding with the taxonomy-guided deep learning method. Latent representations of a patient's visit record and trial criteria can be acquired through the utilization of extensive language models (LLMs). In this work, we employ the pretrained BERT \cite{devlin2018bert} as our text encoder. For patient embedding, a memory network \cite{weston2014memory}, $Mem(\cdot)$, is utilized to manage the patient records, thereby effectively maintaining the sequence of visit data within the embedding space. Formally, the patient record embedding, represented as $x_P$, is derived from the encoding function $f_P(\cdot)$, which can be expressed as:

\begin{align}
    x_P = f_P(P) = Mem(BERT(a_1), BERT(a_2), \ldots, BERT(a_{n}) ),\ a_i \in \{\mathcal{D}, \mathcal{M},\mathcal{P}\}.
\end{align}

In order to capture and encode these essential features within the embedding space, we implement a prior approach that employs a convolutional neural network (CNN) and a highway layer \cite{srivastava2015highway} to extract patterns at various levels for the semantic matching task \cite{you2018end}. Formally, the encoding function $f_c(\cdot)$ is employed to encode an EC embedding $x_c$, which can be described as:

\vspace{-1em}
\begin{equation}
    x_{i/e} = f_c(c) = Highway(BERT(c)),
\end{equation}

where $c \in \mathcal{T}$. The outputs of highway networks are calculated as $
    Highway(\cdot) = Sigmoid(\cdot) Conv(\cdot) + Conv(\cdot) (1 - Sigmoid(Conv(\cdot)))
$.

\textbf{Prediction and Embedding Learning}

Throughout the model optimization process, we aim to maximize patient-trial matching while explicitly addressing the differences between inclusion and exclusion criteria. To achieve this, we design a composite loss function that comprises the following loss terms. The first loss is classification loss. To optimize the classification performance between the predicted outcome $\hat{y}$ and the ground truth $y$, we employ a cross-entropy loss term:

\vspace{-1em}
\begin{equation}
\mathcal{L}_{cla} = -y^T \log(\hat{y}) - (1 - y)^T \log(1 - \hat{y}).
\end{equation}

\begin{figure*}[t!]
\centerline{\includegraphics[width=1\textwidth]{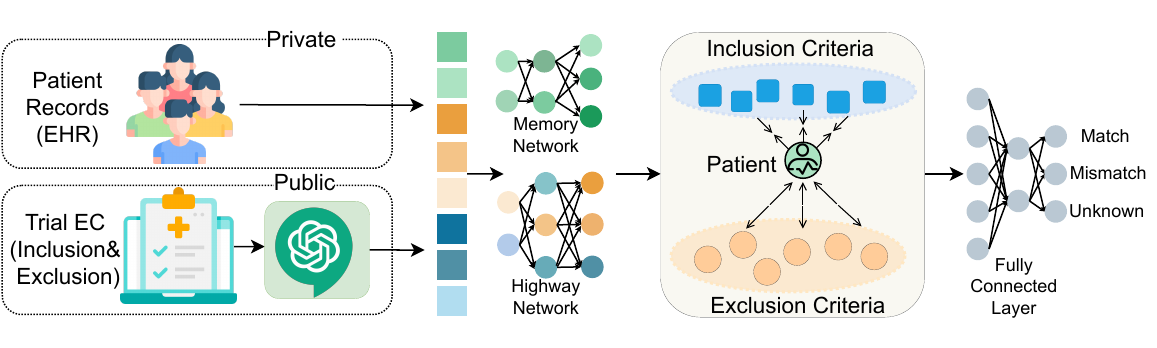}}
\caption{Overall model framework}
\label{fig:framework}
\end{figure*}

Furthermore, we construct inclusion/exclusion contrastive loss term to explicitly address the match between patient embedding and EC embedding for both inclusion and exclusion criteria. This loss term enables the model to extract distinct features (e.g., negation words) within the inclusion/exclusion criteria, thereby aiding the decision to include or exclude a patient. Mathematically, this involves maximizing the similarity between the retrieved patient memory and the embedding for inclusion criteria (i.e., $(x_P, x_i)$) while minimizing the similarity between the memory and the embedding for exclusion criteria (i.e., $(x_P, x_e)$). We formulate the loss term using the following pairwise distance loss:

\vspace{-1em}
\begin{equation}
\mathcal{L}_{con} =
\prod_{a = 1, \ldots, n_i} (1 - s(x_{i_a}, x_P)) \cdot \prod_{b = 1, \ldots, n_e} \max(0, s(x_{e_b}, x_P) - \varepsilon),
\end{equation}

where $s(\cdot, \cdot)$ represents the similarity function between two vectors. In our work, we utilize the cosine similarity function to determine the distance between two data modalities. The hyperparameter $\varepsilon$ signifies the minimum similarity between the exclusion criteria embedding and patient memory. If a patient matches an inclusion criterion, the model maximize the cosine similarity between the two embeddings, making $1 - s(x_i, x_P)$ approach 0. If a patient is excluded due to an exclusion criterion, the similarity between the two embeddings (i.e., $\max(0, s(x_e, x_P) - \varepsilon)$) is minimized and must be no less than $\varepsilon$. This ensures that $x_i$ and $x_e$ have distinct distances to $x_P$ in the latent embedding space. Finally, we minimize the loss functions jointly through backpropagation in an end-to-end manner in Eq.\ref{loss_function},

\vspace{-1em}
\begin{equation}
\mathcal{L} = \alpha \cdot \mathcal{L}_{cla} + (1-\alpha) \cdot \mathcal{L}_{con},
\label{loss_function}
\end{equation}

where $\alpha$ controls the strength of classification loss. The overall framework can be found in Fig.~\ref{fig:framework}.

\section{Experiment}
\vspace{-0.5em}

\subsection{Dataset}
\vspace{-0.5em}

\textbf{Clinical Trial Data.} We collected data from six different stroke clinical trials, namely NCT03735979, NCT03805308, NCT03263117, NCT03496883, NCT03876457, and NCT03545607, using \textit{ClinicalTrials.gov} as our source. Our focus was on both the inclusion and exclusion criteria, resulting in 150 sentence-level statements extracted.

\textbf{Patient EHR Data.} The project was approved by the UTHealth Institutional Review Board (IRB) under HSC-SBMI-21-0529 - ``Re-admission Risk Estimation for Stroke Patients''. Using the stroke patient database, we gathered patient claims data for 825 patients who were enrolled in at least one of the six stroke trials. This data included longitudinal prescription and medical claims data for each patient, encompassing their diagnoses, procedures, and medications. 

\textbf{Augment Data.} In the development of the baseline methods, we utilized open-source libraries to generate augmented data. For our \Algnameabbr{}, which employed a LLM-based prompt augmentation, we incorporated OpenAI's ChatGPT from the version released on March 5th.

We then matched each inclusion/exclusion criteria with its corresponding patient EHR, labeling them as either "match" or "mismatch". To add an element of uncertainty, we also included "unknown" labels for one inclusion criterion and one exclusion criterion randomly selected from another trial, leading to a total of 100,000 labeled pairs.

\subsection{Model and Setting}
\vspace{-0.5em}


\textbf{Model Configuration} In the process of text embedding, we employed the Clinical BERT embeddings as our primary method, as delineated by \cite{alsentzer2019publicly}. This specific model was pre-trained on a substantial dataset comprising 2 million clinical notes, which were extracted from the MIMIC-III v1.4 database \cite{johnson2016mimic}. To optimize the performance of the model, a total of 150,000 iterations were executed during the training phase. The resultant embedding produced by the Clinical BERT model featured a dimensionality of 768. As for the architectural design of the highway network, it consisted of a two-layered convolutional structure with each layer containing 128 channels. This configuration was strategically chosen to enhance the learning capacity of the model while maintaining computational efficiency. The machine learning models are built upon PyTorch.

\textbf{Baselines.}  We benchmarked the proposed data augmentation approach in comparison to original data and the standard data augmentation technique on the dataset. Specifically, we tested 1) Swap Word Augmentation, a method that randomly swaps words within the text and is a subset of the Easy Data Augmentation (EDA) \cite{wei2019eda} proposed by Wei et al; 2) Context Word Augmentation \cite{kobayashi2018contextual, kumar2020data}, which employs BERT to insert contextually relevant words by adding a mask token at a random position in the input text and allowing BERT to predict the appropriate word; 3) Back Translation Augmentation. This technique \cite{sennrich2015improving} involves translating the text to another language and then back to English, yielding a distinct text that maintains the original meaning. In this work, we employed Google Translate in deep translator as a translation framework and utilize German, French, and Spanish as intermediate languages.

\textbf{Training Settings.} In the process of developing our experimental framework, we established our training configurations in accordance with the prevailing settings that are widely accepted within the field. To elaborate, we commenced the training of the entire neural networks from scratch for a predetermined duration of 12 epochs. Throughout this stage, we employed the Adam optimizer \cite{kingma2014adam}, accompanied by a batch size of 128 and a learning rate of 1e-4. The hyperparameter $\alpha$ was set to 0.5 and $\varepsilon$ was set to 0.01.

\textbf{Evaluation Metrics.} The outcomes of our study are presented at two distinct levels: the patient-criteria level and the patient-trial level. To comprehensively evaluate the performance of the model, we employed a variety of widely recognized evaluation metrics, including Precision, Recall, and F1 score.

\subsection{Experimental Results Analysis}
\vspace{-0.5em}

\subsubsection{Overall Performance}
\vspace{-0.5em}

In this section, we aggregate the results from both patient criteria and patient-trial matching across all trials under investigation. A comparative analysis between our proposed methods and the established baseline methods is presented in Tab.~\ref{overall_performance}. The experimental results are obtained by calculating the average of three separate tests. Upon examining the experimental outcomes, we discerned that the implementation of our LLM-based augmentation approach yielded a substantial enhancement in the performance of machine learning models across all evaluation metrics.


\begin{table}[h]
\setlength{\tabcolsep}{7pt}
\centering 
\vspace{5pt}
\caption{Overall performance}
\vspace{-8pt}
\resizebox{\columnwidth}{!}{
\begin{tabular}{lccccccc}
\toprule
\multirow{2}{*}{\textbf{Performance}} &\multicolumn{3}{c}{Patient-Criteria Level} & \multicolumn{3}{c}{Patient-Trial Level}\\
\cmidrule(lr){2-4}\cmidrule(lr){5-7}
& Prec & Rec & F1 & Prec & Rec & F1 \\
\midrule
Vanilla (No Augmentation) & 0.858$\scriptstyle\pm 1.1e-2 $ & 0.841$\scriptstyle\pm 5.7e-3$ & 0.850$\scriptstyle\pm 8.1e-3$ & 0.715$\scriptstyle\pm 2.0e-2$ & 0.748$\scriptstyle\pm 1.7e-2$ & 0.731$\scriptstyle\pm 1.9e-2$ \\ \midrule
Swap Word Augmentation & 0.854$\scriptstyle\pm 1.8e-2$ & 0.814$\scriptstyle\pm 1.6e-2$ & 0.833$\scriptstyle\pm 1.7e-2$ & 0.707$\scriptstyle\pm 1.2e-2$ & 0.759$\scriptstyle\pm 1.4e-2$ & 0.732$\scriptstyle\pm 1.3e-2$ \\ \midrule
Context Word Augmentation & 0.855$\scriptstyle\pm 1.0e-2$ & 0.835$\scriptstyle\pm 1.4e-2$ & 0.844$\scriptstyle\pm 1.2e-2$ & 0.752$\scriptstyle\pm 1.7e-2$ & 0.789$\scriptstyle\pm 1.5e-2$ & 0.770$\scriptstyle\pm 1.6e-2$ \\ \midrule
Back Translation Augmentation & 0.919$\scriptstyle\pm 1.1e-2$ & 0.845$\scriptstyle\pm 1.3e-2$ & 0.880$\scriptstyle\pm 1.2e-2$ & 0.777$\scriptstyle\pm 1.4e-2$ & 0.796$\scriptstyle\pm 1.3e-2$ & 0.787$\scriptstyle\pm 1.3e-2$ \\ \midrule
\textbf{\Algnameabbr{}} & \textbf{0.964$\scriptstyle\pm 8.0e-3$} & \textbf{0.862$\scriptstyle\pm 1.2e-2$} & \textbf{0.910$\scriptstyle\pm 1.0e-2$} & \textbf{0.801$\scriptstyle\pm 1.3e-2$} & \textbf{0.830$\scriptstyle\pm 1.1e-2$} & \textbf{0.815$\scriptstyle\pm 1.1e-2$}\\
\bottomrule
\end{tabular}
}
\label{overall_performance}
\end{table}

More specifically, the average performance improvements for Precision, Recall, and F1 scores at the patient-criteria level are more than $10.6\%$, $2.1\%$, and $6.0\%$, respectively. A more pronounced performance gain is observed at the patient-trial level, with respective improvements of $8.6\%$, $8.2\%$, and $8.4\%$. These performance gains can be attributed to the inclusion of a more diverse dataset (in comparison to the original dataset) and the incorporation of more precise semantic information (relative to other augmentation techniques). In the forthcoming case study, we will delve deeper into the implications and nuances of these findings, providing a more comprehensive understanding of the impact of our proposed LLM-based augmentation method on the performance of machine learning models in this domain.

\subsubsection{Performance Throughout Different Trials} \label{different_trial_performance}
\vspace{-0.5em}

In this section, we aim to evaluate the performance of our proposed model at the patient criteria level across various trials. The primary objective of this analysis is to determine whether criteria in certain trials prove to be more challenging than others and to assess the extent to which the models can achieve satisfactory results. The outcomes of this examination are presented in Tab.~\ref{trials_performance}. Our findings indicate that the performance of the baseline model does indeed falter in some trials, whereas our proposed method demonstrates a more consistent and robust capacity to tackle these challenging tasks. For instance, in Trial 1, the baseline model only managed to attain a precision of $40.2\%$, while our method exhibited an improvement of $24.0\%$. This suggests that our approach, which incorporates enriched and varied semantic information, can significantly enhance the model's performance, particularly when faced with difficult data. In the subsequent case study, we will delve further into the nuances of these observations to better understand the factors contributing to the success of our proposed method.


\begin{table}[h]
\fontsize{8}{10}\selectfont  
\setlength{\tabcolsep}{7pt}
\centering 
\vspace{5pt}
\caption{Performance throughout different trials}
\vspace{-8pt}
\begin{tabular}{lccccccc}
\toprule
\multirow{2}{*}{\textbf{Performance}} & \multicolumn{3}{c}{Vanilla} & \multicolumn{3}{c}{\Algnameabbr{}} \\
\cmidrule(lr){2-4}\cmidrule(lr){5-7}
& Prec & Rec & F1 & Prec & Rec & F1 \\ \midrule
Trial1 (NCT03263117) & 0.402 & 0.382 & 0.391 & 0.642 & 0.502 & 0.563 \\ \midrule
Trial2 (NCT03496883) & 1.000 & 0.925 & 0.961 & 1.000 & 0.976 & 0.987 \\ \midrule
Trial3 (NCT03545607) & 0.788 & 0.749 & 0.768 & 0.882 & 0.801 & 0.839 \\ \midrule
Trial4 (NCT03735979) & 0.601 & 0.557 & 0.578 & 0.633 & 0.557 & 0.592 \\ \midrule
Trial5 (NCT03805308) & 0.915 & 0.902 & 0.908 & 0.933 & 0.900 & 0.916\\ \midrule
Trial6 (NCT03876457) & 0.498 & 0.496 & 0.496 & 0.593 & 0.405 & 0.481 \\
\bottomrule
\end{tabular}
\vspace{-8pt}
\label{trials_performance}
\end{table}

\subsubsection{Generalizability}
\vspace{-0.5em}

 In this section, we focus on assessing the generalizability of our proposed model in comparison to the baseline (vanilla) model. Our objective is to determine the effectiveness of both models when trained on certain trials and subsequently tested on unseen trials. To conduct this analysis, we designed several experimental scenarios: 1) Training on easy trials and testing on easy trials; 2) Training on easy trials and testing on hard trials; 3) Training on mixed trials and testing on hard trials. The level of difficulty is determined by the performance demonstrated in Section~\ref{different_trial_performance}. The results of these experiments are presented in Tab.~\ref{generalize_performance}.


\begin{table}[h]
\fontsize{8}{10}\selectfont  
\setlength{\tabcolsep}{7pt}
\centering
\vspace{5pt}
\caption{Performance of generalizability}
\vspace{-8pt}
\begin{tabular}{lcccccc}
\toprule
\multirow{2}{*}{\textbf{Performance}} & \multicolumn{3}{c}{Vanilla} & \multicolumn{3}{c}{\Algnameabbr{}} \\
\cmidrule(lr){2-4}\cmidrule(lr){5-7}
& Prec & Rec & F1 & Prec & Rec & F1 \\ \midrule
Case 1 & 0.793 & 0.734 & 0.762 & 0.855 & 0.803 & 0.828 \\ \midrule
Case 2 & 0.525 & 0.452 & 0.485 & 0.673 & 0.604 & 0.636 \\ \midrule
Case 3 & 0.566 & 0.535 & 0.550 & 0.742 & 0.656 & 0.696 \\
\bottomrule
\end{tabular}
\label{generalize_performance}
\end{table}

Upon analyzing the outcomes, it becomes evident that the baseline model exhibits weak performance with respect to generalizability. In contrast, our proposed method demonstrates a notable enhancement in accuracy, achieving an average boost of $12.12\%$ across the different scenarios. In an effort to understand the factors contributing to the observed generalizability when training on easy and hard trials, we consider various scenarios. Both the vanilla model and our method perform similarly when training and testing on easy trials, with our method showing a slight accuracy improvement due to its enriched semantic information. When training on easy trials and testing on hard ones, the baseline model struggles to generalize, while our method excels thanks to its strong adaptability using varied semantic information. Even when training on mixed trials, our method outperforms the vanilla model during testing on hard trials, as it can better comprehend the problem domain. Overall, our proposed method demonstrates the potential to enhance machine learning model performance across different contexts and problem domains by efficiently utilizing enriched semantic information.

\subsection{Case Study}
\vspace{-0.5em}

In this section, we present a case study to illustrate the strengths of our proposed method through the use of concrete examples. Specifically, we discuss two examples that demonstrate augmentation's capacity to alleviate challenges posed by hard data and to enrich the semantic information of the given problem domain. By showcasing these examples, we aim to provide a deeper understanding of the benefits of our method in practice.


\textbf{Case 1: Hard Data Easing}

Consider a challenging case where traditional machine learning models struggle to accurately classify patients based on their medical records. In this scenario, our augmentation method generates additional criteria by leveraging the enriched semantic information derived from the language model. These newly created criteria assist the model in capturing nuanced relationships and patterns that may have been overlooked by the baseline model. Consequently, the model is better equipped to handle the complexity of hard data, resulting in improved performance and more accurate classification of patients. As shown in Tab.~\ref{case_study_1}, the initial data cause erroneous predictions, whereas the model trained on augmented data displayed accurate classification. This improved performance indicates that augmented datasets can enhance the efficacy of machine learning models.

\begin{table}[h]
\fontsize{8}{10}\selectfont  
\setlength{\tabcolsep}{7pt}
\centering
\vspace{5pt}
\caption{Case study: ease of hard data. We highlight the clear expressions.}
\vspace{-8pt}
\begin{tabular}{p{0.11\textwidth}|p{0.61\textwidth}|p{0.07\textwidth}|p{0.06\textwidth}}
\toprule
\textbf{Criteria-Level Matching} & \multirow{2}{*}{Criteria} & \multirow{2}{*}{Prediction} & Ground Truth  \\ \midrule
Vanilla & Positive urine or serum pregnancy test for women of child bearing potential. & Match & \multirow{5}{*}{Mismatch}\\
\cmidrule(lr){1-3}
\multirow{2}{*}{\Algnameabbr{}} & \casebox{Women of reproductive age} have received a positive result on their urine or serum pregnancy test. & \multirow{2}{*}{Mismatch} &  \\
  & \casebox{Women who are capable of bearing children} have a positive pregnancy test in their urine or serum. & \multirow{2}{*}{Mismatch} & \\
\bottomrule
\end{tabular}
\label{case_study_1}
\end{table}

\textbf{Case 2: Semantic Enrichment}

In a second example, let us consider a situation where the initial dataset contains limited semantic information, which may hinder the model's ability to make accurate predictions. Our augmentation method addresses this issue by generating a more diverse set of criteria based on the LLM's understanding of the problem domain. By doing so, the method enriches the semantic information available to the model, allowing it to uncover previously undiscovered relationships and patterns within the data. This enriched representation of the problem domain subsequently contributes to better performance. As shown in Tab.~\ref{case_study_2}, the vanilla model exhibited inaccuracies in patient-trial level matching; however, when trained on augmented data, the model demonstrated improved prediction accuracy. This evidence suggests that the incorporation of augmented datasets can enhance the machine learning model's capacity to comprehend specific criteria and establish associations with other related criteria.

\begin{table}[h]
\fontsize{8}{10}\selectfont  
\setlength{\tabcolsep}{7pt}
\centering
\vspace{5pt}
\caption{Case study: semantic enrichment. We highlight the diverse expressions.}
\vspace{-8pt}
\begin{tabular}{p{0.09\textwidth}|p{0.63\textwidth}|p{0.07\textwidth}|p{0.06\textwidth}}
\toprule
\textbf{Trial-Level Matching} & \multirow{2}{*}{Criteria within one trial} & \multirow{2}{*}{Prediction} & Ground Truth  \\ \midrule
Vanilla & Acute ischemic stroke patients. & Mismatch & \multirow{5}{*}{Match}\\
\cmidrule(lr){1-3}
\multirow{4}{*}{\Algnameabbr{}} & Patients suffering from \casebox{a sudden blockage of blood flow} to the brain due to ischemia. & \multirow{4}{*}{Match} &  \\
  & Individuals experiencing \casebox{a sudden onset of neurological deficits} resulting from a lack of blood supply to the brain. &  & \\
  & People with \casebox{an abrupt interruption of blood flow to the brain} caused by an ischemic event. &  & \\
\bottomrule
\end{tabular}
\label{case_study_2}
\end{table}

Through the exploration of these examples, the case study highlights the potential of our proposed method to tackle challenges associated with hard data and enrich semantic information. This, in turn, contributes to a more robust and comprehensive understanding of the problem domain, ultimately leading to improved performance of machine learning models in various contexts.

\section{Related Works}
\vspace{-0.5em}


\textbf{Patient Trial Matching}. Existing patient-trial matching methodologies can be classified into rule-based systems and deep embedding-based models. Rule-based systems endeavor to extract named entities and relationships for trial eligibility criteria (ECs) and devise rules for patient identification. These systems rely either on extensive human annotations such as EliXR \cite{weng2011elixr}, supervised learning classifiers for rule extraction \cite{bustos2018learning}, or a combination of machine learning and rule-based approaches such as Criteria2Query \cite{yuan2019criteria2query} to establish rules for ECs. In recent years, deep embedding-based models, such as DeepEnroll \cite{zhang2020deepenroll}, have emerged. These models jointly embed patient records and trial ECs in a shared latent space, subsequently aligning them using attentive inference. COMPOSE\cite{gao2020compose} represents the state-of-the-art method, it used a fine-grained model design to differentiate between inclusion and exclusion criteria. In this work, we follow COMPOSE as our baseline model structure.

\textbf{Data Augmentation}. In natural language processing (NLP), data augmentation techniques operate at character, word, sentence, and document levels. Character-level methods, such as random insertion, exchange, replacement, or deletion, improve model resilience against textual noise \cite{belinkov2017synthetic, liu2021divaug}. Word-level techniques, like random swap, deletion \cite{wei2019eda}, synonym augmentation \cite{niu2018adversarial}, and word embeddings-based approaches \cite{wang2015s, mrkvsic2016counter}, enhance text classification tasks while maintaining semantic consistency. Contextual augmentation uses MLMs such as BERT \cite{devlin2018bert} and RoBERTa \cite{liu2019roberta} for generating relevant new text \cite{tang2023does}. Sentence and document-level methods include back translation augmentation \cite{sennrich2015improving}. Patient-trial matching faces similar issues in obtaining training data, our proposed method shows that LLMs can help acquire high-quality data.

\section{Conclusion}
\vspace{-0.5em}

In summary, this study has successfully developed and implemented a novel data augmentation method called \Algnameabbr{}, designed to enhance the patient-trial matching process while preserving patient privacy. By leveraging the power of LLMs for data augmentation, our approach demonstrated a significant performance improvement, with a $7.32\%$ increase over the baseline model. 
Moreover, \Algnameabbr{} proved to be highly generalizable, exhibiting a substantial improvement of $12.12\%$ when applied to new datasets. The promising results obtained from the implementation of \Algnameabbr{} highlight the effectiveness of using LLM-empowered data augmentation techniques in the patient-trial matching pipeline. The success of this method has the potential to revolutionize patient-trial matching processes, contribute to improved clinical trial outcomes, and expedite the development of novel therapeutic interventions. The study also paves the way for future research exploring the extension of \Algnameabbr{} to other domains in the healthcare industry, ultimately optimizing data-driven decision-making processes.


\makeatletter
\renewcommand{\@biblabel}[1]{\hfill #1.}
\makeatother

\bibliographystyle{vancouver}
\bibliography{amia}  

\end{document}